\documentclass[runningheads]{llncs}
\usepackage{graphicx}
\usepackage{tabularx}
\usepackage{multirow}
\usepackage{hyperref}
\begin{document}
\newcounter{RQCounter}
\newcommand{\RQ}[2]{
\refstepcounter{RQCounter} \label{#1}
	\textbf{RQ}$_{\arabic{RQCounter}}$.~\emph{#2}
}
\newcommand{\hrq}[1]{\textbf{RQ}$_{\ref{#1}}$}
\newcolumntype{P}[1]{>{\centering\arraybackslash}p{#1}}
\newcolumntype{L}[1]{>{\raggedright\let\newline\\\arraybackslash\hspace{0pt}}m{#1}}
%
%
\title{Prescriptive Process Monitoring: Quo Vadis?}
%
%
\author{Kateryna Kubrak\inst{1} \and
Fredrik Milani\inst{1} \and
Alexander Nolte\inst{1, 2} \and
Marlon Dumas\inst{1}}
\authorrunning{K. Kubrak et al.}
%
\institute{University of Tartu, Tartu, Estonia
\email{\{kateryna.kubrak,milani,alexander.nolte,marlon.dumas\}@ut.ee}\\
\and Carnegie Mellon University, Pittsburgh, PA, USA\\}
\maketitle              
\begin{abstract}
Prescriptive process monitoring methods seek to optimize a business process by recommending interventions at runtime to prevent negative outcomes or poorly performing cases. In recent years, various prescriptive process monitoring methods have been proposed. This paper studies existing methods in this field via a Systematic Literature Review (SLR). In order to structure the field, the paper proposes a framework for characterizing prescriptive process monitoring methods according to their performance objective, performance metrics, intervention types, modeling techniques, data inputs, and intervention policies. The SLR provides insights into challenges and areas for future research that could enhance the usefulness and applicability of prescriptive process monitoring methods. The paper highlights the need to validate existing and new methods in real-world settings, to extend the types of interventions beyond those related to the temporal and cost perspectives, and to design policies that take into account causality and second-order effects.

\end{abstract}
\section{Introduction}
\label{sec:intro}

Process mining is a family of techniques that facilitate the discovery and analysis of business processes based on execution data. Process mining techniques use event logs extracted from enterprise information systems to, for instance, discover process models or to check the conformance of a process with respect to a reference model~\cite{van2012process}. In this setting, an event log is a dataset capturing the execution of a process step by step, including timestamps, activity labels, case identifiers, resources, and other contextual attributes related to each case or each step within a case. 

Over time, the scope of process mining has extended to encompass techniques that predict the outcome of ongoing cases of a process based on machine learning models constructed from event logs~\cite{maggi2014predictive}. Predictions, however, only become useful to users when they are combined with recommendations~\cite{Marquez2018}. In this setting, \textit{prescriptive process monitoring} is a family of methods to recommend interventions during the execution of a case that, if followed, optimize the process with respect to one or more performance indicators~\cite{shoush2021prescriptive}. For instance, an intervention might improve the probability of the desired outcome (e.g.\ on-time delivery) or mitigate negative outcomes (e.g. delivery delays) \cite{metzger2020triggering}. Different prescriptive process monitoring methods have been proposed in the literature. These methods vary in relation to -- among others -- their predictive modeling approach and the interventions they prescribe. In some cases, two different methods aim at achieving the same objective but in different ways. For instance, to avoid an undesired outcome, one method prescribes assigning resources for the next task \cite{SindhgattaGD16}, whereas another recommends the next task to be executed~\cite{LeoniDR20}.

The benefits of prescriptive process monitoring can only be fully realized if these methods prescribe effective interventions and if these prescriptions are followed~\cite{DeesLAR19}. At present, though, the variety and fragmentation of prescriptive monitoring methods makes it difficult to understand which method is likely to be most effective or more likely to be accepted by end users in a given business situation.
There is no overview that captures existing prescriptive monitoring methods, what objectives they pursue, which interventions they prescribe, which data they require, or the extent to which these methods have been validated in real-life settings. 
Research overviews and classification frameworks have been put forward in the related field of predicative monitoring~\cite{di2018predictive,Marquez2018} and automated resource allocation~\cite{arias2018human,DBLP:journals/corr/abs-2107-07264}, but not in the field of prescriptive process monitoring. 

To address this gap, we study three research questions:
\begin{itemize}
    \item Given that prescriptive process monitoring methods aim at prescribing interventions that produce business value, i.e., achieve an objective, we formulate the first research question as \RQ{RQ1}{What is the objective for using prescriptive process monitoring methods in the process?} 
    \item The second research question aims at discovering how the objectives can be achieved: \RQ{RQ2}{What are the interventions prescribed by prescriptive process monitoring methods?}
    \item Finally, we explore the data required by the proposed methods: \RQ{RQ3}{What data do prescriptive process monitoring methods require?}
\end{itemize}

To answer these questions, we conduct a Systematic Literature Review (SLR) following the guidelines proposed by Kitchenham et al.~\cite{kitchenham2007guidelines}. We identified 36 papers that we analyze these papers to develop a multi-dimensional framework to characterize prescriptive monitoring methods. The contribution of the paper is twofold. We first develop a framework that classifies the prescriptive process monitoring methods according to their objective, metric, intervention types, techniques, data inputs, and policies to trigger the interventions. Second, we provide insights into potential areas for future research in this field.

The rest of this paper is structured as follows. Section~\ref{sec:prescriptive} discusses related work. In Section~\ref{sec:method}, we elaborate on the protocol of the SLR, while in Section~\ref{sec:results} we present our findings. In Section~\ref{sec:framework} we present the proposed framework, followed by concluding remarks in Section~\ref{sec:conclusion}.

\section{Related Work}
\label{sec:prescriptive}

Methods for prescriptive process monitoring prescribe interventions that can change the outcomes of an ongoing process case. For instance, if a method detects that an undesired outcome is probable to unfold, an alarm is raised that can lead to an intervention \cite{TeinemaaTLDM18}. This intervention could come in the form of an action performed by a process worker, such as calling a customer, that helps to mitigate or prevent the negative outcome from materializing \cite{TeinemaaTLDM18}. Interventions often entail an intervention cost (e.g., time spent executing an action) and a cost of undesired outcome (e.g., the order is canceled) \cite{fahrenkrog2019fire}. Thus, it is essential to define a policy for when the prescription is generated. The example above \cite{fahrenkrog2019fire} considers the probability of a negative outcome and evaluates the cost model and the mitigation effectiveness before triggering interventions.

A few previous studies focus on areas that are related to prescriptive process monitoring. Di Francescomarino et al.~\cite{di2018predictive} introduce a value-driven framework that allows companies to identify when to apply predictive process monitoring methods. Another classification of predictive process monitoring methods is that of Márquez-Chamorro et al.~\cite{Marquez2018}, where the focus is on methods to train predictive models. In Mertens et al.~\cite{8904693}, the authors evaluate predictive methods used to recommend follow-up activities in the healthcare domain. While prescriptive process monitoring methods incorporate predicted outputs, our work only focuses on prescriptive methods and prescribed interventions.


Pufahl et al.~\cite{DBLP:journals/corr/abs-2107-07264} present an SLR on automatic resource allocation. Similarly, Arias et al.~\cite{arias2018human} give an overview of resource allocation methods, but with a particular focus on human resources. However, the former work reviews prescriptive methods in general, and the latter two focus on a specific intervention, namely, resource allocation. In this paper, we enrich such works by considering all types of potential interventions in process-aware methods.

\section{SLR Method}
\label{sec:method}

We aim to review the existing body of work on prescriptive process monitoring methods. More specifically, what the objectives for using such methods are (\hrq{RQ1}), what interventions the methods prescribe (\hrq{RQ2}), and what data the methods require (\hrq{RQ3}). Therefore, we use the systematic literature review (SLR) method as it aids us to identify relevant literature in a specific research area \cite{kitchenham2007guidelines}. We follow the guidelines proposed by Kitchenham et al.~\cite{kitchenham2007guidelines}, who proposes three main steps: (1) planning the review, (2) conducting it, and (3) reporting the findings. 

For the first step (planning), we identified research questions and developed the review protocol~\cite{kitchenham2007guidelines}. The research questions were defined and motivated above. We developed a search string for the review protocol, identified suitable electronic databases, and defined inclusion and exclusion criteria. Finally, we defined the data extraction strategy. 
In the search string, we included ``process mining'' to scope the study to methods that rely on event logs. We derived the term ``prescriptive'' from the research questions. We also included the terms ``recommender'' and ``decision support'', as we found these to be sometimes used instead of ``prescriptive''. Accordingly, we formulated the search string:

\smallskip
\noindent
\begin{tabular}{|L{12cm}|}
\hline 
\footnotesize
\textit{(recommender OR ``decision support'' OR prescriptive) AND ``process mining''} \\ \hline
\end{tabular}

While conducting the first search, we noted that the term ``prescriptive process monitoring'' is not consistently used. Using this search string only might thus lead us to miss relevant studies. We addressed this by examining the papers identified with the first search string to identify other used terms. We noted that terms such as ``next-step recommendation'', ``next best actions'', ``proactive process adaptation'' are used as synonyms for prescriptive process monitoring. We also noticed that the phrase ``business process'' often appeared in titles and keywords. Therefore, we formulated the second search string as:

\noindent
\begin{tabular}{|L{12cm}|}
\hline
\textit{(recommender OR ``next activity'' OR ``next step'' OR ``next resource'' OR proactive) AND ``business process''} \\ \hline
\end{tabular}

We searched using both strings on ACM Digital Library, Scopus (includes SpringerLink), Web of Science, and IEEE Xplore. The databases were selected based on coverage of publications within the field of process mining. We also ran the search strings on Google Scholar to capture potentially relevant works not yet published (such as arXiv). Finally, we conducted backwards referencing (snowballing)~\cite{okoli2010guide} to identify additional relevant papers.

Next, we defined the exclusion and inclusion criteria. We excluded papers not digitally accessible (EC1), not in English (EC2), duplicates (EC3), and shorter than six pages (EC4). Exclusion criteria EC1 and EC2 ensure that the paper can be generally accessed and understood by other researchers. Papers with open access or accessible via institutional access are considered accessible. 
Criterion EC3 removes duplicates that can appear since several digital libraries are used. We applied criterion EC4, as papers with less than six pages are likely not to contain enough information for our analysis. We also defined three inclusion criteria: (IC1) the paper is relevant to the domain of prescriptive process monitoring, (IC2) the paper presents, reviews, discusses, or demonstrates a method or a case for prescriptive process monitoring, (IC3) the paper describes at least one way to identify candidate interventions for an ongoing process case. IC1 aims to filter out the papers outside of the scope of the prescriptive process monitoring domain. With IC2, the studies that represent any theoretical discussion or practical application of a method are included. Inclusion criterion IC3 ensures that the paper contains enough information to address the research questions.

Finally, we defined the data extraction strategy. We first captured the metadata of all papers (title, author, publication venue, year). Then we defined the data required to address the research questions. Thus, for \hrq{RQ1}, we defined the data required to identify the objective of using the prescriptive process monitoring technique and the performance metric(s) targeted in each of the papers. Next, we defined the data to elicit the interventions prescribed, the process perspective, and the users the prescribed interventions are presented to (\hrq{RQ2}). Finally, we defined the required data input for the techniques described in the different papers (\hrq{RQ3}). Additionally, we added modeling techniques and policies used to trigger interventions.

We ran the search\footnote{The first search was conducted on 22 Sep 2021, the second on 12 Oct 2021.} and identified a total of 1367 papers. We filtered them using the exclusion criteria EC1 and EC2. This resulted in the removal of 97 papers. Thus, 1270 papers remained and were filtered based on EC3. From the remaining 1010 papers, we removed short papers (EC4). This resulted in 900 papers remaining. These were filtered by title, thus removing papers that were clearly out of scope. The remaining 171 papers were filtered by abstract, resulting in 66 papers remaining. Finally, we applied the inclusion criteria by reading the whole paper and removing 44 papers. As a result, 22 papers remained. A total of 14 papers were added through backwards referencing, resulting in the final list of 36 relevant papers\footnote{Full review protocol: \url{https://doi.org/10.6084/m9.figshare.17091554.v3}}.

\begin{table}[]
\centering
\vspace*{-2mm}
\caption{Paper Selection Process}\label{tab:filtering}
\begin{tabular}{L{4cm}|P{1.2cm} P{1.1cm}|P{1.2cm} P{1.1cm} |P{1.2cm} P{1.1cm}}
Search & First & & Second & & Aggregated \\ \hline
Selection criteria & \# found & \# left & \# found & \# left & \# found & \# left \\ \hline
Search results & 572 & & 795 & & 1367 \\
Data cleaning & 60 & 512 & 37 & 758 & 97 & 1270 \\
Filtering by duplicates & 116 & 396 & 144 & 614 & 260 & 1010 \\
Filtering by \# of pages & 31 & 365 & 79 & 535 & 110 & 900 \\
Filtering by paper title & 252 & 113 & 477 & 58 & 729 & 171 \\
Filtering by paper abstract & 64 & 49 & 41 & 17 & 105 & 66 \\
Filtering by full paper & 34 & 15 & 10 & 7 & 44 & 22 \\
Backward referencing & 12 & & 2 & \\
\textbf{Total} & & 27 & & 9 & & \textbf{36} \\
\end{tabular}
\end{table}

To derive the framework, we started with clustering the methods according to what they were aiming to improve (\hrq{RQ1}), e.g., "cycle time minimization", "cost optimization". We then noted that the methods formed two more prominent groups that, in the end, served as the main categorization of the framework, i.e., the objectives. Within the groups, we followed the research questions to classify the methods further, such as according to the interventions they trigger (\hrq{RQ2}), the input data they require (\hrq{RQ3}).

\section{SLR Results}
\label{sec:results}

In the following sections, we present the results of our review. First, we describe the objectives of prescriptive process monitoring methods that we found (\hrq{RQ1}). Then, we present the interventions prescribed to achieve the objectives (\hrq{RQ2}), and, finally, we outline the data used to do so (\hrq{RQ3}).

\subsection{Prescriptive Process Monitoring Objectives}
\label{subsec:objective}

From our review, we identified two main objectives that prescriptive process monitoring methods aim to achieve. The first objective reduces the defect rate, whereas the second relates to optimizing quantitative case performance. The objective of reducing defect rate is expressed with binary metrics. For instance, the objective is achieved by reducing the risk of cost overrun \cite{ConfortiLRAH15}. The objective of optimizing is expressed as, for instance, reducing cycle time \cite{WibisonoNBP15}.

As to the objective of reducing the defect rate, five papers discuss undesired temporal outcomes, such as the violation of the planned cycle time or deadline \cite{groger2014prescriptive,SindhgattaGD16,WeinzierlDZM20,HuberFH15,LeoniDR20}. For instance, Gröger et al. \cite{groger2014prescriptive} describe the example of a manufacturing process, where the target is to avoid exceeding the allowed limits for cycle time. Another set of studies focuses on avoiding or mitigating an undesired categorical outcome \cite{TeinemaaTLDM18,fahrenkrog2019fire,metzger2020triggering,shoush2021prescriptive,ghattas2014improving,thomas2017recommending,Mertens20,HaisjacklW10}. For example, Ghattas et al. \cite{ghattas2014improving} try to avoid the customer rejecting the delivery in a bottle manufacturing process. In the domain of healthcare, examples of undesired outcomes are patients entering a critical stage \cite{thomas2017recommending}, or medical mistakes due to patient restrictions \cite{Mertens20}. One paper aims to eliminate or mitigate process risks, i.e., faults in the process that may arise if not addressed \cite{ConfortiLRAH15}.

The second main objective considers optimizing quantitative case performance. Most papers consider optimizing the temporal perspective (15 out of 20), such as cycle or processing time. More specifically, in \cite{WibisonoNBP15,kim2013constructing,obregon2013dtminer,thomas2017online}, reducing cycle time is defined as the main objective. For instance, Thomas et al. \cite{thomas2017online} describe a method to minimize the cycle time of an environmental permit application process. Others focus on processing time, i.e., time spent by a resource resolving a task \cite{DumasRMR18}. For instance, in Park et al. \cite{ParkS19} the aim is to reduce the processing time of manual tasks in a loan application process. Another set of methods aim at increasing quality. For example, a method seeks to increase perceived service quality for the users of a financial web service \cite{WeinzierlSZM20}. Finally, two papers \cite{GoossensDH18,TerragniH19} describe methods that aim to improve revenues, e.g., by increasing customer lifetime value \cite{GoossensDH18}

\subsection{Prescribed Interventions}
\label{subsec:interventions}

Prescriptive process monitoring methods prescribe actions to take, i.e., interventions. These interventions can be categorized according to the process perspective of the prescribed intervention. Our review indicates that interventions commonly concern control flow and resource perspectives.

A common intervention perspective is control flow, such as prescribing the next task to perform \cite{LeoniDR20,HeberHS15,Batoulis14,nakatumba2012meta}. More specifically, in de Leoni et al. \cite{LeoniDR20}, the next best task is prescribed to the professional who helps a customer in finding a new job, whereas in Weinzierl et al. \cite{WeinzierlSZM20}, the next step is presented to the end-user. Following the prescribed intervention can improve execution time, customer satisfaction or service quality. In other studies, a sequence of next steps is prescribed as an intervention. For instance, in one method, the appropriate treatment of a blood infusion is prescribed for patients based on their personal information \cite{DetroSPLLB20}, whereas another method prescribes steps to be taken in a trauma resuscitation process \cite{YangDSZFXBM17}. Such interventions aim to improve treatment quality.

Another group of methods focuses on the resource perspective, e.g., which resource should perform the next task. For instance, Wibisono et al. \cite{WibisonoNBP15} prescribe which police officer is best suited for the next task in a driving license application process based on their predicted performance. In another method, a mechanic is recommended to carry out car repairs because s/he is predicted to finish it within a defined time given their schedule and experience \cite{SindhgattaGD16}.

Some papers propose prescribing multiple interventions for one case \cite{shoush2021prescriptive,NezhadB11,BarbaWV11}. For example, an intervention to make an offer to a client is prescribed together with a suggestion for a specific clerk to carry out the task \cite{shoush2021prescriptive}. Similarly, in an IT service management process, recommending the next task and the specialist to perform it can help to resolve open cases quicker \cite{NezhadB11}. 

When reviewing the identified papers, we noted that interventions could be divided into two aspects; \textit{intervention frequency} and \textit{intervention basis}. Intervention frequency captures \textit{when} interventions are prescribed. In this sense prescriptive monitoring methods can be \textit{continuous} or \textit{discrete}. If the method is continuous, it prescribes an intervention for multiple or all activities of an ongoing case. For example, the best-suited resource for each next task is prescribed \cite{WibisonoNBP15}. Discrete interventions, in comparison, prescribe actions to be taken only when a need is detected. For instance, in Metzger et al. \cite{metzger2020triggering}, interventions are triggered only when it is detected that the probability of a negative outcome exceeds a defined threshold.

The intervention basis describes whether a method is \textit{prediction}-based or \textit{similarity}-based. Prediction-based methods predict the outcomes of an ongoing case and then prescribe an intervention. Similarity-based methods in comparison provide recommendations solely based on an analysis of historical traces. For instance, one method predicts the possible outcomes if a task is performed as the next step \cite{LeoniDR20}. The next step is prescribed based on which option leads to a greater metric improvement. In contrast to this method, a set of actions are prescribed based on the similarity rate of a current ongoing case and similar previous cases in another method \cite{TrikiSDH13}.

\subsection{Required Data Input}
\label{subsec:data}

Prescriptive process monitoring methods we identified use control flow, resource, temporal, and domain-specific data. Some methods focus on a single type of data, but other methods combine data input from different types. 

As expected, methods that prescribe interventions impacting control flow, such as the next task to execute, commonly use control flow data. For example, in Conforti et al. \cite{ConfortiLRAH15}, the authors apply decision trees on data, such as task duration and frequency, to predict the risk of a case fault, e.g., exceeding the maximum cycle time and costs overrun. Goossens et al. \cite{GoossensDH18} prescribe the next task by using the sequence of events as a key feature.

Data on resources are used to trigger interventions related to different prescription perspectives. For instance, one method predicts the execution time of past resource performance \cite{yaghoibi2017cycle}. The data on resource performance is used to reallocate pending work items to the resources with higher efficiency. In another method, the authors use resource roles and capabilities combined with domain-specific features, such as vehicle type, to recommend which mechanic should be assigned the next task \cite{SindhgattaGD16}. The data is used to predict which resource would improve the probability of the vehicle repair being finished within a defined time. 

Temporal data, e.g.\ day of the week, is also used to prescribe interventions. Such data is commonly used in combination with other data, such as control flow or resources. For instance, the best-suited resource to execute the next task is recommended utilizing the period of the day (morning, afternoon, or evening), inter-arrival rate, and task queue data as input \cite{WibisonoNBP15}. In another method, temporal information (month, weekday, hour) of the last event and the inactive period before the most recent event in the log are used to evaluate the effectiveness of an intervention to reduce cycle time \cite{bozorgi2021prescriptive}.

Domain-specific features, such as materials used in a manufacturing process \cite{ghattas2014improving}, patient demographics and treatment attributes in a patient treatment process \cite{YangDSZFXBM17}, are also utilized as data input. For instance, data about previously treated patients and data on the current patient is used to assess the predicted outcome of alternative next tasks \cite{Mertens20}. This method recommends the task that has the best-predicted outcome for the patient to reduce the risk of medical mistakes, such as prescribing the wrong medication.

\section{Framework and Implications}
\label{sec:framework}

In this section, we present a framework for characterizing existing work on prescriptive process monitoring. Furthermore, we discuss the implications of our review and conclude with the limitations of the study.

\subsection{Framework Overview}

\begin{figure}
\centering
\includegraphics[width=1.6\textwidth, angle = 90]{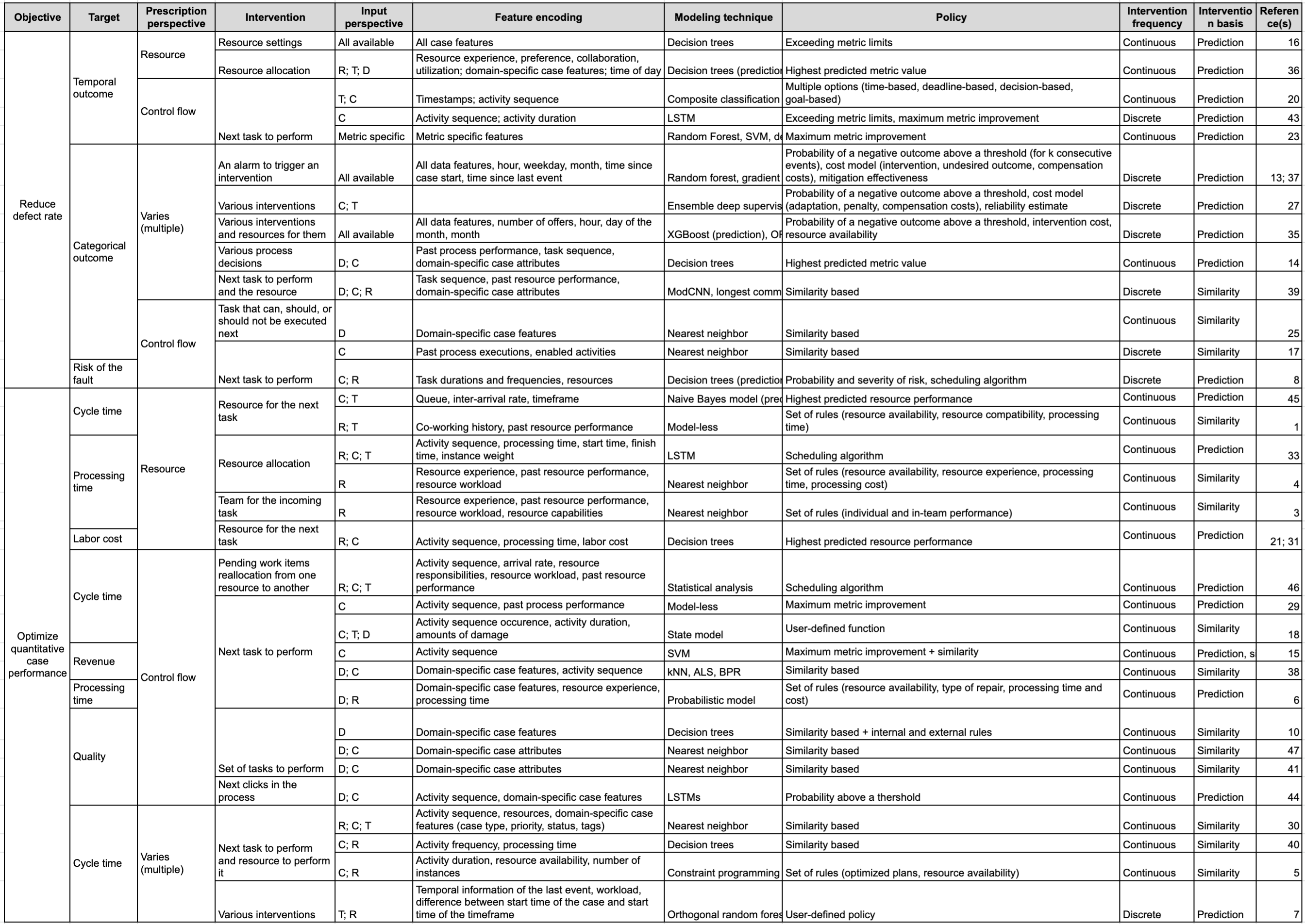}
\caption{Prescriptive process monitoring framework} \label{fig:framework}
\end{figure}

The proposed framework (Figure~\ref{fig:framework}) characterizes prescriptive process monitoring methods from ten aspects, derived from the review results (cf. Section ~\ref{sec:results}). The framework\footnote{Link to the full framework: \url{https://doi.org/10.6084/m9.figshare.17091521.v1}} reads from left to right and begins with the aspect of objective. Other aspects include the interventions to reach the objective, the input required, the techniques used in the methods and the policies used to trigger interventions.

The main aspect of the framework is the \textit{Objective}. Our analysis shows that the identified methods can be divided into two categories according to the objective they pursue (cf. Section ~\ref{subsec:objective}). The first category aims to reduce the percentage of cases with a negative outcome (i.e.\ the defect rate). Methods in the second category aim to optimize a performance dimension captured via a quantitative performance metric defined at the level of each case (e.g.\ cycle time). The next aspect of the framework is the \textit{Target}: the metric used to assess if the performance is improved by the prescribed interventions. For the objective of reducing the defect rate, the target may be a count of a categorical case outcome (e.g., customer complaints) or of a temporal outcome (deadline violations). Quantitative performance targets include cycle time, labor cost, or revenue.

The next two columns (\textit{Prescription perspective, Intervention}) capture the interventions that the methods prescribe to pursue the defined objective. Thus, \textit{Prescription perspective} describes to which process aspect the intervention concerns, e.g., resource or control flow. We also included the category "multiple" for methods that describe several interventions. Then, the aspect \textit{Intervention} lists the actual interventions (cf. section~\ref{subsec:interventions}). For instance, actual intervention can be which resource to assign to the next task.

The following four columns define the data (Input perspective, Feature encoding), techniques (Modeling technique), and policies (Policy) to trigger interventions. Namely, \textit{Input perspective} describes the types of features, i.e., input data, required for a specific method (cf. section~\ref{subsec:data}). Thus, the categories we elicited are (C) control flow (e.g., activities, sequence, and frequencies), (R) resources (e.g., performers of activities), (T) temporal features (time-related), and (D) domain-specific (features that depend on the domain or type of process). The aspect \textit{Feature encoding} explains how features are further refined by a prescriptive method. For instance, resource-perspective features can be encoded as resource experience, resource performance, or resource workload.

The aspect \textit{Modeling technique} relates to the technique used in the method to predict the outcome of the process or the metric performance based on the input. Next, \textit{Policy} relates to the conditions under which an intervention is prescribed. For example, under a similarity-based policy, an intervention is prescribed based on the similarity of the current case to previous cases. Some methods take a set of rules as their policy. For example, the need for an intervention is detected when the probability of a negative outcome exceeds a defined threshold, but also the effectiveness of the intervention is assessed before prescribing it. 

The aspect \textit{Intervention frequency} shows whether a method is continuous (prescribes actions at every step) or discrete (only when the need is detected). Additionally, \textit{Intervention basis} describes whether a method builds on the prediction of how the current case will continue or its similarity to the past cases (cf. section~\ref{subsec:interventions}). Finally, the aspect \textit{Example} (see full version of the framework) can be used as a reference to how the introduced method with its inputs, technique, and policies was used to trigger interventions to reach the objective of a process in a specific domain. 

Existing methods could be explored from the framework by objective, target, and prescribed process perspective. As such, if one seeks to minimize the cycle time of a process, the aim is to optimize quantitative case performance (\textit{Objective}), more specifically, cycle time (\textit{Target}). The framework shows that this can be achieved by prescribing interventions related to the control-flow or resources of the respective process (\textit{Prescription perspective}). If we follow the resource perspective, the framework shows that a set of methods can recommend, for example, resources for the next task or a whole team for a specific request (\textit{Intervention}). For instance, two methods \cite{WibisonoNBP15,abdulhameed2018resource} propose to recommend resources for the next task as an intervention. However, they have different input perspectives and techniques. Continuing with selecting control flow and temporal as the input perspectives, it leads to the method which uses the predicted (\textit{Intervention basis}) highest resource performance (\textit{Policy}) to continuously (\textit{Intervention frequency}) prescribe a resource for the next task.

\subsection{Research Gaps and Implications}

The presented framework provides an overview of existing prescriptive process monitoring methods by categorizing them according to their objectives and targets. The framework also presents different methods available and different ways these methods enable reaching particular objectives. The overview unveils several gaps and associated implications for future research. 

First, we observe that in the vast majority of previous studies, the proposed prescriptive process monitoring methods have been tested using synthetic and/or real-life event logs. However, the validation of the method is done using a real-world or synthetic \textit{observational} event log, but not in real-life settings. An attempt to test the effectiveness of interventions in real-life settings was made by Dees et al. \cite{DeesLAR19}. Their study showed that the predictions were rather accurate, but the interventions did not lead to the desired outcomes. Thus, validations of methods should be done in real-life settings to ensure their usefulness in practice, as also previously highlighted by \cite{Marquez2018}.

Second, the current state-of-the-art in the field is focused on identifying cases in which interventions should be applied and finding the point in time an intervention should be triggered during the execution of a case. In contrast, little attention has been given to the problem of discovering which interventions could be applied to optimize a process with respect to a performance objective. Discrete methods leave it up to the users (stakeholders) to define the possible intervention(s) a priori (e.g., \cite{LeoniDR20,TeinemaaTLDM18}). Those methods that use observational event logs from BPI challenges\footnote{\url{https://www.tf-pm.org/competitions-awards/bpi-challenge}}, rely on winner reports to identify the possible interventions (e.g., \cite{shoush2021prescriptive,TeinemaaTLDM18}). Continuous methods, on the other hand, focus on recommending the next task(s) (e.g., \cite{GoossensDH18,Batoulis14,YangDSZFXBM17}) or resource allocation (e.g., \cite{AriasMS16,abdulhameed2018resource,kim2013constructing}). Thus, one direction for further research is designing methods to support the discovery of interventions from business process event logs, or from textual documentation, or other unstructured or structured metadata about the process. 

Related to the above problem of discovering interventions, another gap in existing research relates to the problem of designing and tuning policies for prescriptive process monitoring. Existing prediction-based methods (e.g., \cite{SindhgattaGD16,groger2014prescriptive}) prescribe an intervention when the probability of a negative outcome exceeds a defined threshold. However, because the predictive models upon which methods rely are based on correlation (as opposed to causal relations), the prescriptions produced by this method might not address the cause of the negative outcome or poor performance (e.g.\ the cause of delay). In this respect, we note that only a few existing methods take into account causality in policy design (e.g., \cite{shoush2021prescriptive,bozorgi2021prescriptive}). Thus, developing policy design techniques that take into account causality is another direction for further research.

As discussed in~\cite{DeesLAR19,LeoniDR20}, the choice of whether or not to apply an intervention or which intervention to apply often depends on contextual factors. Some interventions may lead to prove ineffective or counter-productive, for example, due to second-order effects. For example, an intervention wherein a customer is contacted pro-actively in order to prevent a customer complaint may actually heighten the probability of a complaint~\cite{DeesLAR19}. Similarly, assigning a resource to a case that is running late might lead to other cases being neglected, thus creating delays elsewhere and thus leading to a higher ratio of delayed cases. Detecting such second-order effects requires human judgment and iterative policy validation (e.g.\ via AB testing). In this respect, it is striking that existing prescriptive process monitoring methods do not take into account the need to interact with human decision-makers. A crucial step in this direction is the ability to explain why the prescriptive monitoring system recommends a given intervention. There are two aspects to this question. First, explaining the prediction that underpins a given prescription (prediction explanation), and second, explaining the policy that is used to trigger the prescription (policy explanation). A possible direction to enhance the applicability of prescriptive monitoring methods in practice is to integrate explainability mechanisms into them. While several proposals have been made to enhancement the explainability of predictive process monitoring methods~\cite{HsiehMO21}, the question of policy explanation in the area of prescriptive process monitoring is unexplored. In other words, current methods do not incorporate a mechanism to explain why an intervention is recommended by the method for a given case and in a given state.

Besides the above gaps, the SLR highlights that the majority of methods in the field aim to improve processes along the temporal perspective (e.g.\ cycle time, processing time, deadline violations). In comparison, other performance dimensions are represented in only a few examples (quality in \cite{DetroSPLLB20,WeinzierlSZM20,YangDSZFXBM17}, revenue in \cite{GoossensDH18,TerragniH19}). Thus, another research direction is to investigate other performance objectives that could be enhanced via prescriptive process monitoring.

Finally, our review also highlights a lack of common terminology in the field. This might be due to the novelty of the research field of prescriptive process monitoring. The literature refers to methods with a wide range of names. As such, the terms "proactive process adaptation", "on-the-fly resource allocation", "next-step recommendation" are all used to describe the development and application of prescriptive process monitoring methods. This highlights the need for common terminology. 

\subsection{Threats to Validity}

SLRs have a number of typical pitfalls and threads to validity~\cite{kitchenham2007guidelines,AMPATZOGLOU2019201}. First, there is a potential risk of missing relevant publications during the search. We mitigated this risk by conducting a two-phase search that included a broad range of key terms, as well as backward referencing. Another potential threat is to exclude relevant publications during screening. We mitigated this threat by using explicitly defined inclusion and exclusion criteria. Furthermore, all unclear cases were examined and discussed by all authors of this paper. Third, there is a threat of data extraction bias as this step involves a degree of subjectivity. We discussed each paper in the final list and refined the data extraction when needed to minimize this risk.

\section{Conclusion}
\label{sec:conclusion}

This paper provided a snapshot of the research field of prescriptive process monitoring via an SLR and outlined a framework for characterizing methods in this field. The framework characterizes existing methods according to their objective, target metric, intervention type, technique, data input, and policy used to trigger interventions. The framework was derived from and used to characterize the 36 relevant studies identified by the SLR.


Based on the SLR, we identified research gaps and associated research avenues. In particular, the SLR highlighted: (i) a lack of \emph{in vivo} validation of the proposed methods; (ii) a lack of methods for discovering suitable interventions and assessing their potential effectiveness; (iii) little emphasis on explainability and feedback loops between the prescriptive monitoring system and the end-users; and (iv) a narrow focus on temporal metrics and comparatively little work on applying prescriptive monitoring to other performance dimensions.


\smallskip\noindent\textbf{Acknowledgments.} This research is funded by the Estonian Research Council (PRG1226) and the European Research Council (PIX Project).
\vspace*{-2mm}

\bibliographystyle{splncs04}
\bibliography{references}
\end{document}